\documentclass[twocolumn,conference]{IEEEtran}
\usepackage[T1]{fontenc}
\usepackage[latin9]{inputenc}
\usepackage{color}
\usepackage{float}
\usepackage{booktabs}
\usepackage{amsmath}
\usepackage{amsthm}
\usepackage{amssymb}
\usepackage{graphicx}
\usepackage{setspace}
\usepackage[unicode=true,
 bookmarks=true,bookmarksnumbered=true,bookmarksopen=true,bookmarksopenlevel=1,
 breaklinks=false,pdfborder={0 0 0},pdfborderstyle={},backref=false,colorlinks=false]
 {hyperref}
\hypersetup{pdftitle={Your Title},
 pdfauthor={Your Name},
 pdfpagelayout=OneColumn, pdfnewwindow=true, pdfstartview=XYZ, plainpages=false}

\makeatletter

\providecommand{\tabularnewline}{\\}
\floatstyle{ruled}
\newfloat{algorithm}{tbp}{loa}
\providecommand{\algorithmname}{Algorithm}
\floatname{algorithm}{\protect\algorithmname}

\theoremstyle{plain}
\newtheorem{thm}{\protect\theoremname}

\usepackage{algorithmic}
\usepackage{algorithm}
\usepackage[font=small]{caption}

\@ifundefined{showcaptionsetup}{}{%
 \PassOptionsToPackage{caption=false}{subfig}}
\usepackage{subfig}
\makeatother

\providecommand{\theoremname}{Theorem}

\begin{document}
\title{Bayesian Federated Model Compression for Communication and Computation
Efficiency}
\author{\IEEEauthorblockN{Chengyu~Xia\IEEEauthorrefmark{1}, Danny H. K. Tsang\IEEEauthorrefmark{1}\IEEEauthorrefmark{2},
and Vincent K. N. Lau\IEEEauthorrefmark{1}}\IEEEauthorblockA{\IEEEauthorrefmark{1}Department of Electronic and Computer Engineering,
The Hong Kong University of \\
Science and Technology, Kowloon, Hong Kong}\IEEEauthorblockA{\IEEEauthorrefmark{2}Internet of Things Thrust, The Hong Kong University
of Science and Technology (Guangzhou), Guangzhou, China\\
Email: \{\href{mailto:cxiaab@connect.ust.hk}{cxiaab@connect.ust.hk},
\href{mailto:eetsang@ust.hk}{eetsang@ust.hk}, \href{mailto:eeknlau@ece.ust.hk}{eeknlau@ece.ust.hk}\}}}
\maketitle
\begin{abstract}
In this paper, we investigate Bayesian model compression in federated
learning (FL) to construct sparse models that can achieve both communication
and computation efficiencies. We propose a decentralized Turbo variational
Bayesian inference (D-Turbo-VBI) FL framework where we firstly propose
a hierarchical sparse prior to promote a clustered sparse structure
in the weight matrix. Then, by carefully integrating message passing
and VBI with a decentralized turbo framework, we propose the D-Turbo-VBI
algorithm which can (i) reduce both upstream and downstream communication
overhead during federated training, and (ii) reduce the computational
complexity during local inference. Additionally, we establish the
convergence property for thr proposed D-Turbo-VBI algorithm. Simulation
results show the significant gain of our proposed algorithm over the
baselines in reducing communication overhead during federated training
and computational complexity of final model. 
\end{abstract}

\IEEEpeerreviewmaketitle{}

\section{Introduction}

Recently, federated learning (FL) has emerged as a promising technique
for collaborative learning with distributed data \cite{zhang2022personalized,Chor2023,McMahan2017,Odeyomi2023}.
Although FL has proven to be highly efficient in harnessing the distributed
data, it is still faced with several challenges: \textbf{\emph{1)
High communication overhead during training: }}Since modern deep neural
networks (DNNs) usually have a massive number of weights, uploading
and downloading the models can induce huge communication overhead
between the server and clients. \textbf{\emph{2) Heavy computation
burden during local inference:}} The large DNN models can also induce
heavy computation burden during local inference because of the limited
computation resources at the clients.{\let\thefootnote\relax\footnotetext{This work was supported in part by the Hong Kong Research Grants Council under the Areas of Excellence Scheme Grant AoE/E-601/22-R, in part by the Hong Kong Research Grants Council under Grant 16207221, in part by Guangzhou Municipal Science and Technology Project under Grant 2023A03J0011, in part by Guangdong Provincial Key Laboratory of Integrated Communications, Sensing and Computation for Ubiquitous Internet of Things, and in part by National Foreign Expert Project under Project Number G2022030026L.}}

In order to address the first issue, there are many researches focused
on communication efficient FL. In \cite{Ji2022}, the local models
are top-k sparsified before each uploading. In \cite{louizos2021federated},
the local models are sparsified by training with a spike-slab prior.
In \cite{reisizadeh2020fedpaq,xu2020ternary}, the local models are
quantized to reduce the uploading bits. However, in most existing
works, the local models are compressed \emph{independently} at each
client, and usually have\emph{ different sparse structures}. This
can cause \emph{the aggregated model to lose sparsity} and hence,
the downstream communication overhead as well as the final model computational
complexity are both not reduced \cite{shah2021model}.

In parallel, model compression has been investigated to address the
second issue in a \emph{centralized} scenario. The key idea is to
formulate the model training as a deterministic problem with some
regularization assigned to the weights \cite{zhuang2020neuron,yang2020harmonious,shen2022prune},
or a Bayesian inference problem with some sparse promoting prior on
the weights \cite{Louizos2017,van2020bayesian}, such that the unimportant
weights can be forced to zero during training. However, existing model
compression approaches are not computation efficient enough as they
cannot achieve a \emph{regular enough} sparse structure \cite{xia2023structured,kwon2020structured}.
Moreover, since they are designed for a centralized scenario, they
\emph{cannot} be directly applied in FL scenario where the dataset
is decentralized \cite{Kairouz2021}.

In this paper, we propose a decentralized Turbo variational Bayesian
inference (D-Turbo-VBI) FL framework, which, to our best knowledge,
is the first time to achieve \emph{both communication and computation
efficiency }in FL from the \emph{Bayesian} perspective. The following
summarizes our contributions. 

\textbf{Bayesian model compression with hidden Markov model (HMM)
based hierarchical sparse prior:} We firstly design an HMM based hierarchical
sparse prior for the DNN weights. The proposed prior can induce a
clustered sparse structure in the weight matrix, which is highly efficient
in communication and computation. We further formulate the federated
model compression as a Bayesian inference problem which aims to infer
the weights posterior given the hierarchical prior and the decentralized
dataset.

\textbf{Decentralized Turbo-VBI algorithm for FL:} To solve the Bayesian
inference problem, we propose a decentralized Turbo variational Bayesian
inference (D-Turbo-VBI) algorithm which combines \emph{message passing}
and \emph{VBI} with a \emph{decentralized turbo framework}. By careful
design of the message passing, the proposed algorithm can 1) capture
a clustered sparse structure based on the uploaded messages from clients
and the hierarchical prior, and 2) promote a \emph{common} sparse
structure among all local models. With all local models possessing
the common structure, both upstream and downstream communication overhead
as well as the computational complexity of the final model can be
reduced. Moreover, convergence property for the D-Turbo-VBI algorithm
is also established.

\section{Proposed Sparse Structure and Problem Formulation}

\subsection{Proposed Clustered Sparse Structure}

Although existing model compression methods can zero out entire neurons
\cite{Neill2020,scardapane2017group}, the surviving weights are still
randomly distributed, which makes the transmission and computation
of these weight matrix still need to be performed in element wise
\cite{kwon2020structured,yu2017scalpel}. Thus, we propose a novel
clustered sparse structure where the non-zero elements tend to gather
in \emph{clusters} in the weight matrix, as shown in Fig. \ref{clustered_structure}a.
Compared to existing sparse structures, our proposed clustered sparse
structure has the following advantages: 1) weight matrix can be coded
and transmitted in cluster wise, which can greatly reduce the communication
overhead. 2) Acceleration techniques such as matrix tiling can be
applied to support cluster-wise matrix multiplication, which can greatly
reduce the computational complexity \cite{kwon2020structured}.

\subsection{HMM Based Hierarchical Prior}

In order to promote the clustered structure, we design a HMM based
hierarchical prior for the DNN weights. Consider a DNN model with
$L$ layers and $N$ weights. Let $w$ denote a weight and each $w$
is associated with a support $s\in\left\{ 0,1\right\} $. Denote the
precision of $w$ by $\rho$. If $s=0$, the corresponding $\rho$
satisfies $\mathbb{E}\left[\rho\right]\gg1$ such that the corresponding
$w$ has a very small prior variance. Thus, the value of $w$ is forced
to stay close to its mean. If $s=1$, the corresponding $\rho$ satisfies
$\mathbb{E}\left[\rho\right]=\mathcal{O}\left(1\right)$. Thus, the
corresponding $w$ has more freedom to take different values. Based
on the above relationship, the hierarchical prior is given as
\begin{equation}
p\left(\mathbf{w},\boldsymbol{\rho},\mathbf{s}\right)=p\left(\mathbf{s}\right)p\left(\boldsymbol{\rho}|\mathbf{s}\right)p\left(\mathbf{w}|\boldsymbol{\rho}\right),\label{eq:prior}
\end{equation}
 where $\mathbf{w}$ denotes all the DNN weights, $\boldsymbol{\rho}$
denotes the precisions, and $\mathbf{s}$ denotes the supports. The
detailed probability models for $p\left(\mathbf{s}\right)$, $p\left(\boldsymbol{\rho}|\mathbf{s}\right)$
and $p\left(\mathbf{w}|\boldsymbol{\rho}\right)$ are specified below.

\begin{figure}[tbh]
\begin{centering}
\subfloat[]{\begin{centering}
\includegraphics[width=0.17\textwidth]{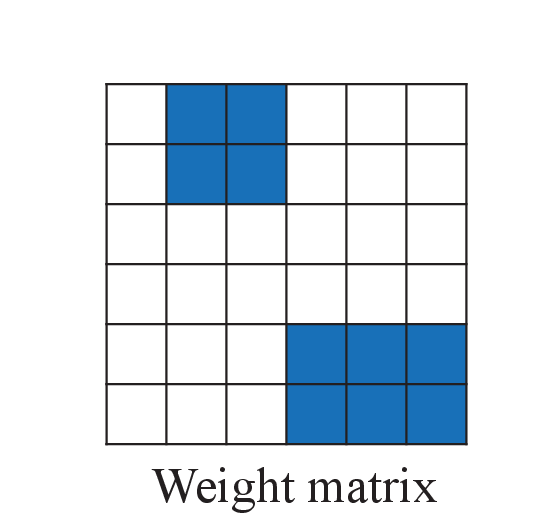}
\par\end{centering}
}\hfill{}\subfloat[]{\begin{centering}
\includegraphics[width=0.25\textwidth]{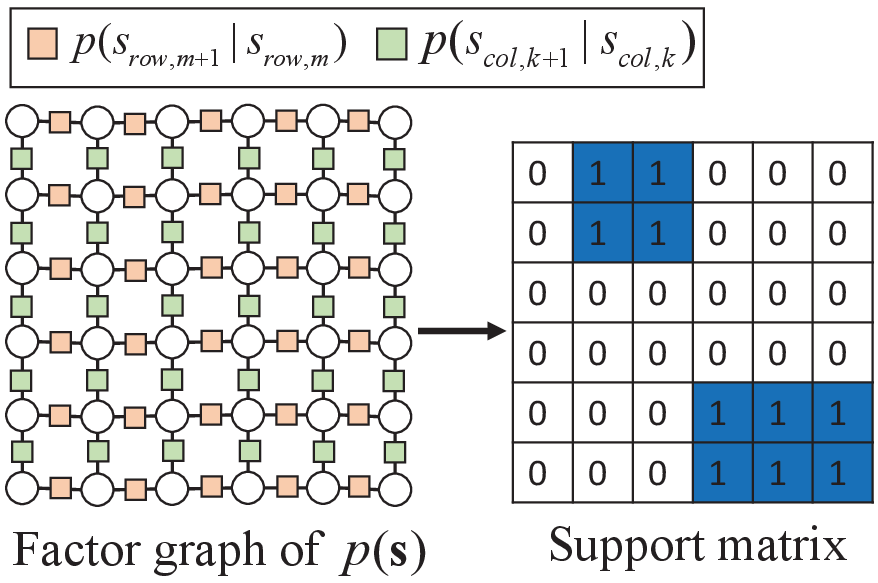}
\par\end{centering}
}
\par\end{centering}
\centering{}\caption{(a) Proposed clustered sparse structure on a $6\times6$ weight matrix.
Non-zero elements are marked in blue. (b) HMM prior for $p\left(\mathbf{s}\right)$
on a $6\times6$ weight matrix. Non-zero supports will gather in clusters
due to the correlation between neighbors.\label{clustered_structure}}
\end{figure}

\textbf{\emph{Probability Model for $p\left(\mathbf{s}\right)$:}}
Let $\mathbf{s}_{l}$ denote the supports in the $l$-th layer, $p\left(\mathbf{s}\right)$
can be factorized in layer wise as $p\left(\mathbf{s}\right)=\prod_{l=1}^{L}p\left(\mathbf{s}_{l}\right)$.
Assume the $l$-th layer has a $K\times M$ weight matrix, \textbf{\emph{$p\left(\mathbf{s}_{l}\right)$}}
is modeled as an HMM with a $K\times M$ grid structure, as shown
in Fig. \ref{clustered_structure}b. In particular, the HMM is parameterized
by transition probabilities $p\left(s_{row,m+1}|s_{row,m}\right)$
and $p\left(s_{col,k+1}|s_{col,k}\right)$, where $s_{row,m}$ is
the $m$-th support in the row vector and $s_{col,k}$ is the $k$-th
support in the column vector. The average cluster size in the weight
matrix can be adjusted by tuning the values of the transition probabilities.

\textbf{\emph{Probability Model for $p\left(\boldsymbol{\rho}|\mathbf{s}\right)$:}}\emph{
}$p\left(\boldsymbol{\rho}|\mathbf{s}\right)$ is modeled as a Bernoulli
Gamma distribution governed by the value of $\mathbf{s}$:
\begin{equation}
p\left(\boldsymbol{\rho}|\mathbf{s}\right)=\prod_{n=1}^{N}\left(\Gamma\left(\rho_{n};a_{n},b_{n}\right)\right)^{s_{n}}\left(\Gamma\left(\rho_{n};\overline{a}_{n},\overline{b}_{n}\right)\right)^{1-s_{n}},
\end{equation}
 Specifically, the parameters satisfy $\frac{a_{n}}{b_{n}}=\mathbb{E}\left[\rho_{n}|s_{n}=1\right]=\mathcal{O}\left(1\right)$
and $\frac{\overline{a}_{n}}{\overline{b}_{n}}=\mathbb{E}\left[\rho_{n}|s_{n}=0\right]\gg1$.
We choose $p\left(\boldsymbol{\rho}|\mathbf{s}\right)$ as the Gamma
distribution due to its conjugacy with the Gaussian distribution,
which allows for a closed-form solution in the later algorithm design.

\textbf{\emph{Probability Model for $p\left(\mathbf{w}|\boldsymbol{\rho}\right)$:}}
$p\left(\mathbf{w}|\boldsymbol{\rho}\right)$ is modeled as a Gaussian
distribution which is commonly used \cite{liu2021bayesian,zhang2022personalized}:
\begin{equation}
p\left(\mathbf{w}|\boldsymbol{\rho}\right)=\prod_{n=1}^{N}p\left(w_{n}|\rho_{n}\right),\quad p\left(w_{n}|\rho_{n}\right)=\mathcal{N}\left(w_{n};0,\frac{1}{\rho_{n}}\right).
\end{equation}

\subsection{Federated Model Compression Formulation}

Let $\mathcal{D}=\left\{ \left(x_{1},y_{1}\right),\left(x_{2},y_{2}\right),...,\left(x_{d},y_{d}\right),...,\left(x_{\left|\mathcal{D}\right|},y_{\left|\mathcal{D}\right|}\right)\right\} $
denote the global dataset where $\left(x_{d},y_{d}\right)$ is the
$d$-th feature-label pair and $\left|\mathcal{D}\right|$ is the
size of the dataset. Let $T$ denote the total client number in the
FL system, the global dataset $\mathcal{D}$ is distributed across
the $T$ clients and each client possesses a local dataset $\mathcal{D}_{t}$,
and we have $\mathcal{D}=\cup_{t=1}^{T}\mathcal{D}_{t}$. 

Let $\Lambda=\left\{ \mathbf{w},\boldsymbol{\rho},\mathbf{s}\right\} $
denote the variable set for simplicity. With the proposed prior $p\left(\Lambda\right)$
in \eqref{eq:prior}, the primary goal of Bayesian FL is to calculate
the global model posterior $p\left(\Lambda|\mathcal{D}\right)$. Based
on the Bayes theorem, the posterior $p\left(\Lambda|\mathcal{D}\right)$
is given as 
\begin{equation}
p\left(\Lambda|\mathcal{D}\right)=\frac{p\left(\mathcal{D}|\Lambda\right)p\left(\Lambda\right)}{p\left(\mathcal{D}\right)},\label{eq:posterior}
\end{equation}
where
\begin{equation}
p\left(\mathcal{D}|\Lambda\right)=\prod_{d=1}^{\left|\mathcal{D}\right|}p\left(y_{d}|x_{d},\Lambda\right),
\end{equation}
 is the likelihood of the neural network model. However, directly
calculating $p\left(\Lambda|\mathcal{D}\right)$ is intractable even
with a much simpler prior in a centralized setting \cite{Louizos2017,Jospin2020}.
In the following, we propose a D-Turbo-VBI algorithm to calculate
the posterior in an FL setting. 

\section{Decentralized Turbo-VBI Algorithm}

\subsection{Decentralized Turbo Framework}

\begin{figure}[t]
\begin{centering}
\includegraphics[width=0.45\textwidth]{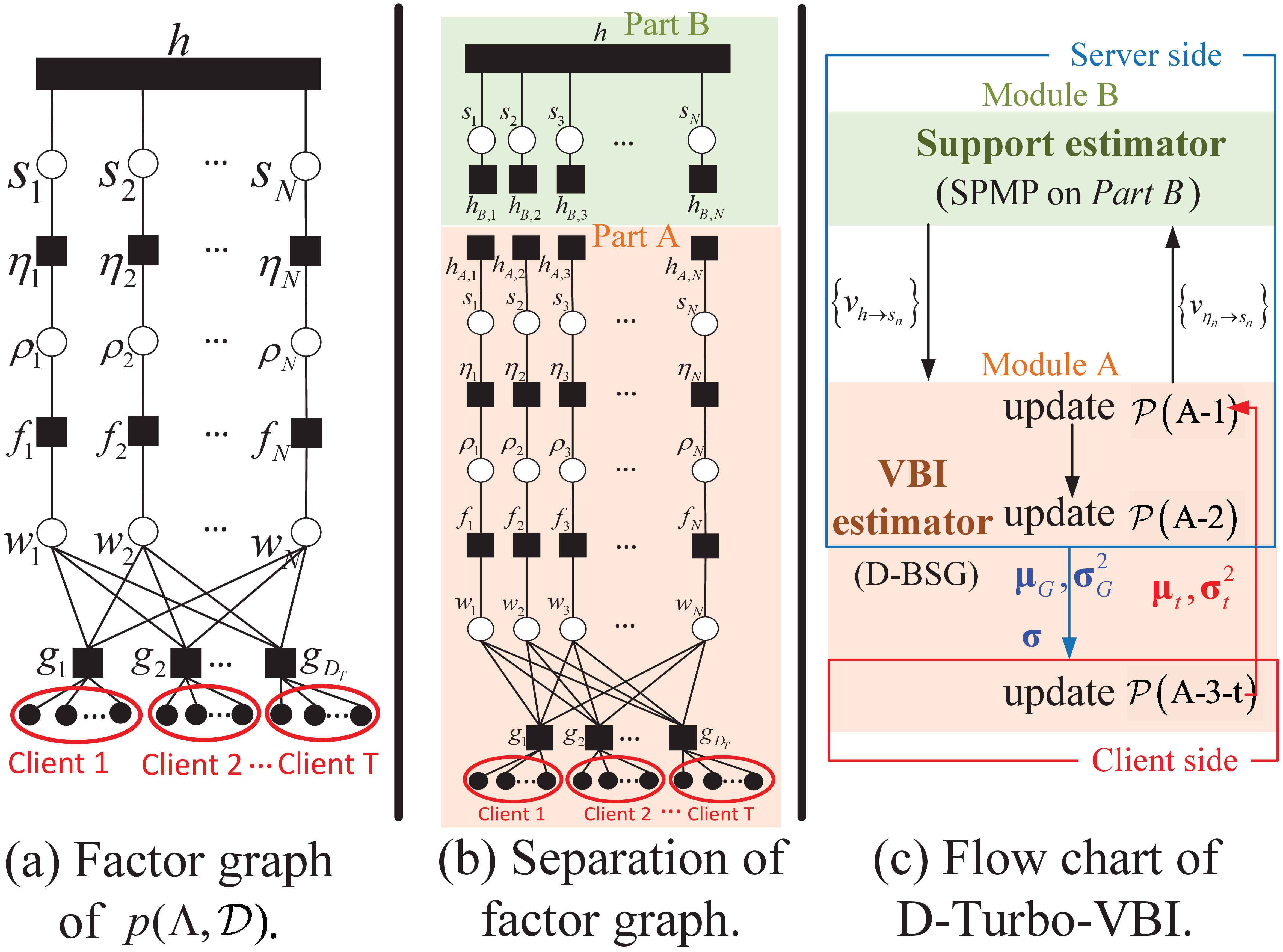}
\par\end{centering}
\centering{}\caption{Illustration of the decentralized Turbo-VBI FL framework. \label{fig:D-Turbo-VBI chart}}
\end{figure}

Our proposed D-Turbo-VBI algorithm stems from message passing over
the factor graph of the joint distribution $p\left(\Lambda,\mathcal{D}\right)$.
The factor graph of $p\left(\Lambda,\mathcal{D}\right)$ is plotted
in Fig. \ref{fig:D-Turbo-VBI chart}a where $g$ denotes the likelihood
function of the neural network $p\left(y_{d}|x_{d},\mathbf{w}\right)$,
$f$ denotes the weight prior $p\left(w_{n}|\rho_{n}\right)$, $\eta$
denotes the precision prior $p\left(\rho_{n}|s_{n}\right)$, and $h$
denotes the support prior $p\left(\mathbf{s}\right)$. In order to
handle the HMM prior $p\left(\mathbf{s}\right)$, we propose to separate
the factor graph into two parts following the turbo framework \cite{liu2020robust}:
\emph{Part B} which contains the HMM prior $p\left(\mathbf{s}\right)$
and \emph{Part A} which contains the remaining part, as shown in Fig.
\ref{fig:D-Turbo-VBI chart}b. Correspondingly, our algorithm has
two modules: \emph{Module B} performs sum product message passing
(SPMP) in \emph{Part B} to obtain a structured support prior, and
send the output message $v_{h\rightarrow s_{n}}$ to \emph{Module
A. }While\emph{ Module A} performs VBI to obtain an approximate posterior
$q\left(\Lambda\right)$ of the original $p\left(\Lambda|\mathcal{D}\right)$,
and the output message $v_{\eta_{n}\rightarrow s_{n}}$ is given by
$\frac{q\left(s_{n}\right)}{v_{h\rightarrow s_{n}}}$. The two modules
exchange messages iteratively until convergence according to the message
passing rule. 

\subsection{Design of Varitaional Posterior}

Since we perform VBI in \emph{Module A}, the first task is to design
a tractable variational posterior $q\left(\Lambda\right)$. For simplicity,
mean-field assumption \cite{Louizos2017,tzikas2008variational} is
adopted: $q\left(\Lambda\right)=\prod_{n=1}^{N}q\left(w_{n}\right)q\left(\rho_{n}\right)q\left(s_{n}\right).$
In the following, we elaborate on the designs of $q\left(w_{n}\right)$,
$q\left(\rho_{n}\right)$ and $q\left(s_{n}\right)$, respectively.

\textbf{\emph{Design of $q\left(w_{n}\right)$:}} With the prior $p\left(w_{n}|\rho_{n}\right)$
being a Gaussian distribution, it is straightforward to model $q\left(w_{n}\right)$
as a Gaussian distribution:
\begin{equation}
q\left(w_{n};\mu_{n},\sigma_{n}^{2}\right)=\mathcal{N}\left(w_{n};\mu_{n},\sigma_{n}^{2}\right),
\end{equation}
 where $\mu_{n}$ and $\sigma_{n}$ are the variational posterior
mean and standard deviation, respectively.

\textbf{\emph{Design of $q\left(\rho_{n}\right)$:}} With $p\left(w_{n}|\rho_{n}\right)$
being a Gaussian distribution, the prior $p\left(\rho_{n}\right)$
is conjugate to a Gamma distribution. Thus, $q\left(\rho_{n}\right)$
is modeled as a Gamma distribution:
\begin{equation}
q\left(\rho_{n};\widetilde{a}_{n},\widetilde{b}_{n}\right)=\Gamma\left(\rho_{n};\widetilde{a}_{n},\widetilde{b}_{n}\right),
\end{equation}
 where $\widetilde{a}_{n}$ and $\widetilde{b}_{n}$ are the variational
parameters.

\textbf{\emph{Design of $q\left(s_{n}\right)$:}} With $\hat{p}\left(\mathbf{s}\right)$
being a Bernoulli distribution, it is straightforward to model $q\left(s_{n}\right)$
as a Bernoulli distribution:
\begin{equation}
q\left(s_{n};\widetilde{\pi}_{n}\right)=\left(\widetilde{\pi}_{n}\right)^{s_{n}}\left(1-\widetilde{\pi}_{n}\right)^{1-s_{n}},
\end{equation}
 where the active probability $\widetilde{\pi}_{n}$ is the variational
parameter.

We summarize the variational parameter set by $\xi=\left\{ \boldsymbol{\mu},\boldsymbol{\sigma},\widetilde{\boldsymbol{a}},\widetilde{\boldsymbol{b}},\widetilde{\boldsymbol{\pi}}\right\} $. 

\subsection{Module A}

\begin{singlespace}
With $p\left(\mathbf{s}\right)$ separated into \emph{Part B}, the
support prior in \emph{Part A} naturally becomes an i.i.d. Bernoulli
distribution given as
\begin{equation}
\hat{p}\left(\mathbf{s}\right)=\prod_{n=1}^{N}\left(\pi_{n}\right)^{s_{n}}\left(1-\pi_{n}\right)^{1-s_{n}},
\end{equation}
where $\pi_{n}$ is the active probability, given by
\begin{equation}
\pi_{n}=p\left(s_{n}=1\right)=\frac{v_{h\rightarrow s_{n}}\left(1\right)}{v_{h\rightarrow s_{n}}\left(1\right)+v_{h\rightarrow s_{n}}\left(0\right)}.
\end{equation}
With $\hat{p}\left(\mathbf{s}\right)$, we have a new joint prior
$\hat{p}\left(\Lambda\right)$ in \emph{Module A}, which is given
as:
\begin{equation}
\hat{p}\left(\Lambda\right)=\hat{p}\left(\mathbf{s}\right)p\left(\boldsymbol{\rho}|\mathbf{s}\right)p\left(\mathbf{w}|\boldsymbol{\rho}\right).\label{eq:new prior}
\end{equation}
 The main task of \emph{Module A} is to achieve the optimal variational
posterior $q^{\star}\left(\Lambda;\xi\right)$ with the new prior
$\hat{p}\left(\Lambda\right)$, which is realized by minimizing the
negative evidence lower bound (ELBO) \cite{Jospin2020}:

\begin{equation}
\mathcal{P}\left(\text{A}\right):\qquad\begin{alignedat}{1}q^{\star}\left(\Lambda;\xi\right)=\arg\min_{\xi} & f_{\text{NELB}}\left(\xi\right)\end{alignedat}
,\label{eq:local vbi problem}
\end{equation}
 where $f_{\text{NELB}}\left(\xi\right)=D_{KL}\left(q\left(\Lambda;\xi\right)||\hat{p}\left(\Lambda\right)\right)-\mathbb{E}_{q\left(\Lambda;\xi\right)}\ln p\left(\mathcal{D}|\Lambda\right)$
is the negative ELBO. 

To solve $\mathcal{P}\left(\text{A}\right)$, we propose a decentralized
block stochastic gradient (D-BSG) algorithm in \emph{Module A}, where
we separate the variational parameter set $\xi$ into three blocks
$\left\{ \widetilde{\boldsymbol{a}},\widetilde{\boldsymbol{b}}\right\} $,
$\widetilde{\boldsymbol{\pi}}$ and $\left\{ \boldsymbol{\mu},\boldsymbol{\sigma}\right\} $,
and iteratively update each block. In the following, we specify the
detailed updating rules for each block.
\end{singlespace}

\begin{singlespace}
\noindent \textbf{\emph{$\bullet$ Updates of $\widetilde{\boldsymbol{\pi}}$:
}}Let $\widetilde{\boldsymbol{a}}=\widetilde{\boldsymbol{a}}^{\left(r\right)}$,
$\widetilde{\boldsymbol{b}}=\widetilde{\boldsymbol{b}}^{\left(r\right)}$,
$\boldsymbol{\mu}=\boldsymbol{\mu}^{\left(r\right)}$, and $\boldsymbol{\sigma}=\boldsymbol{\sigma}^{\left(r\right)}$,
where the superscript $r$ indicates the solutions in the $r$-th
iteration. After dropping all the irrelevant constant terms, the subproblem
for updating $\widetilde{\boldsymbol{\pi}}$ is given by
\begin{equation}
\mathcal{P}\left(\text{A-1}\right):\qquad\min_{\widetilde{\boldsymbol{\pi}}}\mathcal{R}_{\mathbf{s}}\left(\widetilde{\boldsymbol{\pi}}\right),
\end{equation}
where 
\[
\mathcal{R}_{\mathbf{s}}\left(\widetilde{\boldsymbol{\pi}}\right)=D_{KL}\left(q\left(\mathbf{s};\widetilde{\boldsymbol{\pi}}\right)||\exp\left(\left\langle \ln p\left(\boldsymbol{\rho}|\mathbf{s}\right)\hat{p}\left(\mathbf{s}\right)\right\rangle _{q\left(\boldsymbol{\rho}\right)}\right)\right),
\]
and $\left\langle f\left(x\right)\right\rangle _{q\left(x\right)}=\int f\left(x\right)q\left(x\right)dx$.
The optimal solution for $\mathcal{P}\left(\text{A-1}\right)$ can
be achieved with a closed form \cite{liu2020robust}:
\begin{equation}
\widetilde{\pi}_{n}^{\left(r+1\right)}=\frac{C_{1}^{\left(r\right)}}{C_{1}^{\left(r\right)}+C_{2}^{\left(r\right)}}.
\end{equation}
 Specifically, $C_{1}^{\left(r\right)}$ and $C_{2}^{\left(r\right)}$
are given by
\begin{equation}
\begin{aligned}C_{1}^{\left(r\right)} & =\frac{\pi_{n}b_{n}^{a_{n}}}{\Gamma\left(a_{n}\right)}e^{\left(a_{n}-1\right)\left\langle \ln\rho_{n}\right\rangle ^{\left(r\right)}-b_{n}\left\langle \rho_{n}\right\rangle ^{\left(r\right)}},\\
C_{2}^{\left(r\right)} & =\frac{\left(1-\pi_{n}\right)\overline{b}_{n}^{\overline{a}_{n}}}{\Gamma\left(\overline{a}_{n}\right)}e^{\left(\overline{a}_{n}-1\right)\left\langle \ln\rho_{n}\right\rangle ^{\left(r\right)}-\overline{b}_{n}\left\langle \rho_{n}\right\rangle ^{\left(r\right)}},
\end{aligned}
\end{equation}
where $\left\langle \ln\rho_{n}\right\rangle =\psi\left(\widetilde{a}_{n}\right)-\ln\left(\widetilde{b}_{n}\right)$
and $\psi\left(x\right)=\frac{d}{dx}\ln\left(\Gamma\left(x\right)\right)$
is the digamma function.
\end{singlespace}

\noindent \textbf{\emph{$\bullet$ Updates of $\left\{ \widetilde{\boldsymbol{a}},\widetilde{\boldsymbol{b}}\right\} $:
}}Let $\widetilde{\boldsymbol{\pi}}=\widetilde{\boldsymbol{\pi}}^{\left(r+1\right)}$,
$\boldsymbol{\mu}=\boldsymbol{\mu}^{\left(r\right)}$, and $\boldsymbol{\sigma}=\boldsymbol{\sigma}^{\left(r\right)}$,
the subproblem for updating $\left\{ \widetilde{\boldsymbol{a}},\widetilde{\boldsymbol{b}}\right\} $
is given as follows after dropping all the irrelevant constant terms
\begin{equation}
\mathcal{P}\left(\text{A-2}\right):\qquad\min_{\widetilde{\boldsymbol{a}},\widetilde{\boldsymbol{b}}}\mathcal{R}_{\boldsymbol{\rho}}\left(\widetilde{\boldsymbol{a}},\widetilde{\boldsymbol{b}}\right),
\end{equation}
where
\[
\begin{aligned} & \mathcal{R}_{\boldsymbol{\rho}}\left(\widetilde{\boldsymbol{a}},\widetilde{\boldsymbol{b}}\right)\\
 & =D_{KL}\left(q\left(\boldsymbol{\rho};\widetilde{\boldsymbol{a}},\widetilde{\boldsymbol{b}}\right)||\exp\left(\left\langle \ln p\left(\mathbf{w}|\boldsymbol{\rho}\right)p\left(\boldsymbol{\rho}|\mathbf{s}\right)\right\rangle _{q\left(\mathbf{s}\right)q\left(\mathbf{w}\right)}\right)\right).
\end{aligned}
\]
The optimal solution for $\mathcal{P}\left(\text{A-2}\right)$ can
also be achieved with a closed form \cite{liu2020robust}:
\begin{equation}
\begin{aligned}\widetilde{a}_{n}^{\left(r+1\right)} & =\widetilde{\pi}_{n}^{\left(r+1\right)}a_{n}+\left(1-\widetilde{\pi}_{n}^{\left(r+1\right)}\right)\overline{a}_{n}+1,\\
\widetilde{b}_{n}^{\left(r+1\right)} & =|\mu_{n}^{\left(r\right)}|^{2}+\left(\sigma_{n}^{\left(r\right)}\right)^{2}+\widetilde{\pi}_{n}^{\left(r+1\right)}b_{n}+\left(1-\widetilde{\pi}_{n}^{\left(r+1\right)}\right)\overline{b}_{n}.
\end{aligned}
\end{equation}
\textbf{\emph{$\bullet$ Updates of $\left\{ \boldsymbol{\mu},\boldsymbol{\sigma}^{2}\right\} $:
}}Let $\widetilde{\boldsymbol{\pi}}=\widetilde{\boldsymbol{\pi}}^{\left(r+1\right)}$,
$\widetilde{\boldsymbol{a}}=\widetilde{\boldsymbol{a}}^{\left(r+1\right)}$
and $\widetilde{\boldsymbol{b}}=\widetilde{\boldsymbol{b}}^{\left(r+1\right)}$,
after dropping all the irrelevant constant terms, the subproblem for
updating $\boldsymbol{\mu}$ and $\boldsymbol{\sigma}^{2}$ is given
by
\begin{equation}
\mathcal{P}\left(\text{A-3}\right):\qquad\begin{aligned}\min_{\boldsymbol{\mu},\boldsymbol{\sigma}} & \mathcal{L}\left(\boldsymbol{\mu},\boldsymbol{\sigma};\mathcal{D}\right),\end{aligned}
\end{equation}
where 
\begin{equation}
\begin{aligned}\begin{aligned}\mathcal{L}\left(\boldsymbol{\mu},\boldsymbol{\sigma};\mathcal{D}\right) & =\end{aligned}
 & \sum_{n=1}^{N}\left(\ln\frac{\widetilde{\sigma}_{n}^{\left(r+1\right)}}{\sigma_{n}}+\frac{\sigma_{n}^{2}+\mu_{n}^{2}}{2\left(\widetilde{\sigma}_{n}^{\left(r+1\right)}\right)^{2}}-\frac{1}{2}\right)\\
 & -\mathbb{E}_{q\left(\mathbf{w}\right)}\ln p\left(\mathcal{D}|\mathbf{w}\right),
\end{aligned}
\label{eq:fl_global_loss}
\end{equation}
 and $\widetilde{\sigma}_{n}^{\left(r+1\right)}=\sqrt{\frac{\widetilde{b_{n}}^{\left(r+1\right)}}{\widetilde{a_{n}}^{\left(r+1\right)}}}$.
As the global dataset $\mathcal{D}$ is decentralized, \eqref{eq:fl_global_loss}
can be further decomposed into each client as
\begin{equation}
\mathcal{L}\left(\boldsymbol{\mu},\boldsymbol{\sigma};\mathcal{D}\right)=\sum_{t}p\left(\left(x_{d},y_{d}\right)\in\mathcal{D}_{t}\right)\cdot\mathcal{L}_{t}\left(\boldsymbol{\mu},\boldsymbol{\sigma};\mathcal{D}_{t}\right),
\end{equation}
 where
\begin{equation}
\begin{aligned}\mathcal{L}_{t}\left(\boldsymbol{\mu},\boldsymbol{\sigma};\mathcal{D}_{t}\right)= & \sum_{n=1}^{N}\left(\ln\frac{\widetilde{\sigma}_{n}^{\left(r+1\right)}}{\sigma_{n}}+\frac{\sigma_{n}^{2}+\mu_{n}^{2}}{2\left(\widetilde{\sigma}_{n}^{\left(r+1\right)}\right)^{2}}-\frac{1}{2}\right)\\
 & -\frac{1}{p\left(\left(x_{d},y_{d}\right)\in\mathcal{D}_{t}\right)}\cdot\mathbb{E}_{q\left(\mathbf{w}\right)}\ln p\left(\mathcal{D}_{t}|\mathbf{w}\right),
\end{aligned}
\end{equation}
is the local loss function. Thus, each client only needs to update
its local posterior $q_{t}\left(\mathbf{w}\right)$ by minimizing
$\mathcal{L}_{t}\left(\boldsymbol{\mu},\boldsymbol{\sigma};\mathcal{D}_{t}\right)$:
\begin{equation}
\mathcal{P}\left(\text{A-3-t}\right):\qquad\begin{aligned}\min_{\boldsymbol{\mu}_{t},\boldsymbol{\sigma}_{t}} & \mathcal{L}_{t}\left(\boldsymbol{\mu}_{t},\boldsymbol{\sigma}_{t};\mathcal{D}_{t}\right),\end{aligned}
\end{equation}
 and upload the updated $\boldsymbol{\mu}_{t},\boldsymbol{\sigma}_{t}^{2}$
to the server. Then, the server can generate the global posterior
$q_{G}\left(\mathbf{w}\right)$ by aggregating the local parameters
$\boldsymbol{\mu}_{t}$ and $\boldsymbol{\sigma}_{t}^{2}$ . To further
reduce the communication overhead, we let the weights at each client
have the same variational posterior variance: $\boldsymbol{\sigma}_{t}^{2}=\sigma_{t}^{2}\cdot\mathbf{1}$
for $t=1,2,...,T$. Specifically, the aggregations for $\boldsymbol{\mu}_{t}$
and $\boldsymbol{\sigma}_{t}$ are given by 

\noindent 
\begin{equation}
\begin{alignedat}{1}\boldsymbol{\mu}_{G} & =\sum_{t}p\left(\left(x_{d},y_{d}\right)\in\mathcal{D}_{t}\right)\cdot\boldsymbol{\mu}_{t},\\
\boldsymbol{\sigma}_{G} & =\sum_{t}p\left(\left(x_{d},y_{d}\right)\in\mathcal{D}_{t}\right)\cdot\boldsymbol{\sigma}_{t}.
\end{alignedat}
\label{eq:aggregation}
\end{equation}
After the server obtains $q_{G}\left(\mathbf{w}\right)$, it can update
$q\left(\boldsymbol{\rho}\right)$ and $q\left(\mathbf{s}\right)$
for the next iteration. Then, it can calculate the message $v_{\eta_{n}\rightarrow s_{n}}=\frac{q\left(s_{n}\right)}{v_{h\rightarrow s_{n}}}$
from \emph{Module A} to \emph{Module B}.

\subsection{Module B}

With the input message $v_{\eta_{n}\rightarrow s_{n}}$ from \emph{Module
A}, the server performs standard SPMP on the factor graph of \emph{Part
B}. Specifically, the factor graph of \emph{Part B} is given in Fig.
\ref{clustered_structure}b.

\subsection{Overall Algorithm and Convergence Analysis}

The overall D-Turbo-VBI FL algorithm is summarized in Algorithm 1
and a flow chart is provided in Fig. \ref{fig:D-Turbo-VBI chart}c.
\begin{algorithm}[tbh]
\begin{algorithmic}[1] \small
\renewcommand{\algorithmicrequire}{\textbf{Input:}} 
\renewcommand{\algorithmicensure}{\textbf{Output:}}
\REQUIRE Local dataset \{$\mathcal{D}_t$\}, prior parameters $a$, $b$, $\overline a$, $\overline b$, $p\left( {{s_{row,m + 1}}|{s_{row,m}}} \right)$, $p\left( {{s_{col,k + 1}}|{s_{col,k}}} \right)$, number of communication rounds $I_{\text{glb}}$, number of local SGD steps $I_{\text{lcl}}$. 
\ENSURE Variational parameter set $\xi^\star$.
\STATE{\textbf{Initialize:} $\boldsymbol{\mu}_{G}^{r}$, $\boldsymbol{\sigma}_{G}^{r}$, $\widetilde{a}_{n},\widetilde{b}_{n},\pi_{n},\widetilde{\pi}_{n}$.} 
\FOR{$r=0,1,2,...,I_{\text{glb}}$} 
\STATE{\textbf{Client Side:}} 
\STATE{\textbf{Download} $\boldsymbol{\mu}_{t}^{0}\leftarrow\boldsymbol{\mu}_{G}^{r}$, $\boldsymbol{\sigma}_{t}^{0}\leftarrow\boldsymbol{\sigma}_{G}^{r}$, $\widetilde{\boldsymbol{\sigma}}=\sqrt{\frac{\widetilde{\boldsymbol{b}}}{\widetilde{\boldsymbol{a}}}}$.} 
\FOR{$l=0,1,2,...,I_{\text{lcl}}$} 
\STATE $\boldsymbol{\mu}_{t}^{l+1}\leftarrow\boldsymbol{\mu}_{t}^{l}-\eta^{\left(l\right)}\cdot\nabla_{\boldsymbol{\mu}}\mathcal{L}_{t}$ 
\STATE $\boldsymbol{\sigma}_{t}^{l+1}\leftarrow\boldsymbol{\sigma}_{t}^{l}-\eta^{\left(l\right)}\cdot\nabla_{\boldsymbol{\sigma}}\mathcal{L}_{t}$ 
\ENDFOR 
\STATE{\textbf{Upload} $\boldsymbol{\mu}_{t}^{I_{\text{lcl}}+1}$, $\boldsymbol{\sigma}_{t}^{I_{\text{lcl}}+1}$} 
\STATE{\textbf{Server Side:}}

\STATE{Update $q\left(\mathbf{s}\right)$ by (14).} 

\STATE{Update $q\left(\boldsymbol{\rho}\right)$ by (17).} 

\STATE{$\boldsymbol{\mu}_{G}^{r+1}\leftarrow\sum_{t}p\left(\left(x_{d},y_{d}\right)\in\mathcal{D}_{t}\right)\cdot\boldsymbol{\mu}_{t}^{I_{\text{lcl}}+1}$} 
\STATE{$\boldsymbol{\sigma}_{G}^{r+1}\leftarrow\sum_{t}p\left(\left(x_{d},y_{d}\right)\in\mathcal{D}_{t}\right)\cdot\boldsymbol{\sigma}_{t}^{I_{\text{lcl}}+1}$} 

\STATE{Calculate and send message $v_{\eta_{n}\rightarrow s_{n}}$ to \textit{Module B}.} 

\STATE{Perform SPMP on \textit{Module B}, send message $v_{h\rightarrow s_{n}}$ to \textit{Module A}.} 

\ENDFOR 
\end{algorithmic}

{\footnotesize{}\caption{Decentralized Turbo-VBI FL framework}
}{\footnotesize\par}

\end{algorithm}

The convergence of the proposed D-Turbo-VBI algorithm essentially
relies on the convergence of the D-BSG algorithm in \emph{Module A}
as the convergence of SPMP in \emph{Module B} and the message passing
between two modules are quite standard.\textcolor{blue}{{} }The D-BSG
algorithm in \emph{Module A} can be regarded as a combination of block
coordinate descent \cite{tseng2001convergence} and SGD within federated
learning framework \cite{yu2019parallel,xu2015block}. For simplicity,
let $\xi_{1}=\left\{ \widetilde{\boldsymbol{a}},\widetilde{\boldsymbol{b}}\right\} $,
$\xi_{2}=\widetilde{\boldsymbol{\pi}}$, $\xi_{3}=$ $\left\{ \boldsymbol{\mu},\boldsymbol{\sigma}\right\} $
and $\xi_{3,t}=\left\{ \boldsymbol{\mu}_{t},\boldsymbol{\sigma}_{t}\right\} $
where subscript $t$ denotes the $t$-th client. Following \cite{yu2019parallel,xu2015block},
we first make the following assumptions:

\noindent \textbf{Assumption 1.} \emph{(Boundedness and smoothness
of $f_{\text{NELB}}$)} The global objective function $f_{\text{NELB}}\left(\xi_{1},\xi_{2},\xi_{3}\right)$
is lower bounded, i.e. $f_{\text{NELB}}\left(\xi_{1},\xi_{2},\xi_{3}\right)>-\infty$,
and smooth with modulus $L_{1}$.

\noindent \textbf{Assumption 2.} \emph{(Smoothness of $\mathcal{L}_{t}\left(\xi_{3};\mathcal{D}_{t}\right)$)}
Each local loss function $\mathcal{L}_{t}\left(\xi_{3};\mathcal{D}_{t}\right)$
is smooth with modulus $L_{2}$.

\noindent \textbf{Assumption 3.} \emph{(Boundedness of local stochastic
gradient)} Let $\widetilde{g}_{t}$ denote the stochastic gradient
of $\mathcal{L}_{t}\left(\xi_{3,t};\mathcal{D}_{t}\right)$ during
local SGD, and $g_{t}=\nabla\mathcal{L}_{t}\left(\xi_{3,t};\mathcal{D}_{t}\right)$
denote the true gradient. There exists constant $\sigma_{g}>0,$$A>0$
and $G>0$ such that 
\begin{equation}
\mathbb{E}\left[\left\Vert \widetilde{g}_{t}-g_{t}\right\Vert ^{2}\right]\leq\sigma_{g}^{2},\;\left\Vert \mathbb{E}\left[\widetilde{g}_{t}-g_{t}\right]\right\Vert \leq A,\;\mathbb{E}\left[\left\Vert \widetilde{g}_{t}\right\Vert ^{2}\right]\leq G^{2},\:\forall t.
\end{equation}

\noindent \textbf{Assumption 4.} \emph{(Boundedness of variational
parameters $\xi$)} Let $l$ denote the local SGD iteration index.
There exists a constant $\rho>0$ such that $\mathbb{E}\left[\left\Vert \xi^{\left(l\right)}\right\Vert ^{2}\right]\leq\rho^{2}$
for all $l$. 

With the above assumptions, we have the following convergence theorem: 
\begin{thm}
\begin{singlespace}
(Convergence of D-BSG algorithm in Module A)\emph{ Let $\xi_{1}=\left\{ \widetilde{\boldsymbol{a}},\widetilde{\boldsymbol{b}}\right\} $,
$\xi_{2}=\widetilde{\boldsymbol{\pi}}$, $\xi_{3}=$ $\left\{ \boldsymbol{\mu},\boldsymbol{\sigma}\right\} $,
and if the local SGD step size $\eta^{\left(l\right)}$ satisf}ies\emph{
\begin{equation}
\sum_{l=1}^{\infty}\eta^{\left(l\right)}=+\infty,\qquad\sum_{l=1}^{\infty}\left(\eta^{\left(l\right)}\right)^{2}<+\infty,
\end{equation}
where $l$ is the iteration index of local SGD. Then, any cluster
point of the sequence $\{\xi_{1}^{(r+1)},\xi_{2}^{(r+1)},\xi_{3}^{(r)}\}$
generated by the D-BSG algorithm in }Module A\emph{ is a stationary
point of $f_{\text{NELB}}\left(\xi\right)$ where $r$ is the communication
round index.}
\end{singlespace}
\end{thm}
\begin{IEEEproof}
\begin{singlespace}
Due to page limit, we only provide proof outline here. For detailed
proof, please see appendix. Let $\overline{\xi}_{3}^{\left(l\right)}=\sum_{t}p\left(\left(x_{d},y_{d}\right)\in\mathcal{D}_{t}\right)\cdot\xi_{3,t}^{\left(l\right)}$,
the proof mainly consists of the following steps: 1) Prove $||\nabla_{\overline{\xi}_{3}}f_{\text{NELB}}(\xi_{1}^{\left(l+1\right)},\xi_{2}^{\left(l+1\right)},\overline{\xi}_{3}^{\left(l\right)})||$
converges to zero with $l$. 2) Since the sequence $\{\xi_{1}^{\left(r+1\right)},\xi_{2}^{\left(r+1\right)},\xi_{3}^{\left(r\right)}\}$
is a subsequence of $\{\xi_{1}^{\left(l+1\right)},\xi_{2}^{\left(l+1\right)},\overline{\xi}_{3}^{\left(l\right)}\}$,
we have $||\nabla_{\xi_{3}}f_{\text{NELB}}(\xi_{1}^{\left(r+1\right)},\xi_{2}^{\left(r+1\right)},\xi_{3}^{\left(r\right)})||$
converges to zero with $r$. 3) Consider any subsequence $\{\xi_{1}^{\left(s+1\right)},\xi_{2}^{\left(s+1\right)},\xi_{3}^{\left(s\right)}\}$
of $\{\xi_{1}^{\left(r+1\right)},\xi_{2}^{\left(r+1\right)},\xi_{3}^{\left(r\right)}\}$
which converges to some cluster point, prove $||\nabla_{\xi_{1}}f_{\text{NELB}}(\xi_{1}^{\left(s+1\right)},\xi_{2}^{\left(s+1\right)},\xi_{3}^{\left(s\right)})||$
and $||\nabla_{\xi_{2}}f_{\text{NELB}}(\xi_{1}^{\left(s+1\right)},\xi_{2}^{\left(s+1\right)},\xi_{3}^{\left(s\right)})||$
converge to zero with $s$.
\end{singlespace}
\end{IEEEproof}

\section{Performance Analysis}

In this section, we evaluate the performance of the proposed D-Turbo-VBI
algorithm. We experiment on CIFAR-10 dataset with a downsized Alexnet
and CIFAR-100 with MobileNetV1. The client number is set to 10. To
align with real-life scenarios, in both experiments we use \emph{non-i.i.d.}
local datasets generated by a Dirichlet distribution with $\alpha=0.5$
\cite{hsu2019measuring}. For our algorithm, a cluster larger than
$3\times3$ is regarded as an entirety. The transmission bit length
is set to 16 bits for all methods. We choose \emph{Dynamic Sampling
and Selective Masking (DSSM) }\cite{Ji2022} and \emph{FL with Periodic
Averaging and Quantization (FedPAQ) }\cite{reisizadeh2020fedpaq}
as the baselines.
\begin{figure}[tbh]
\begin{centering}
\includegraphics[width=0.47\textwidth]{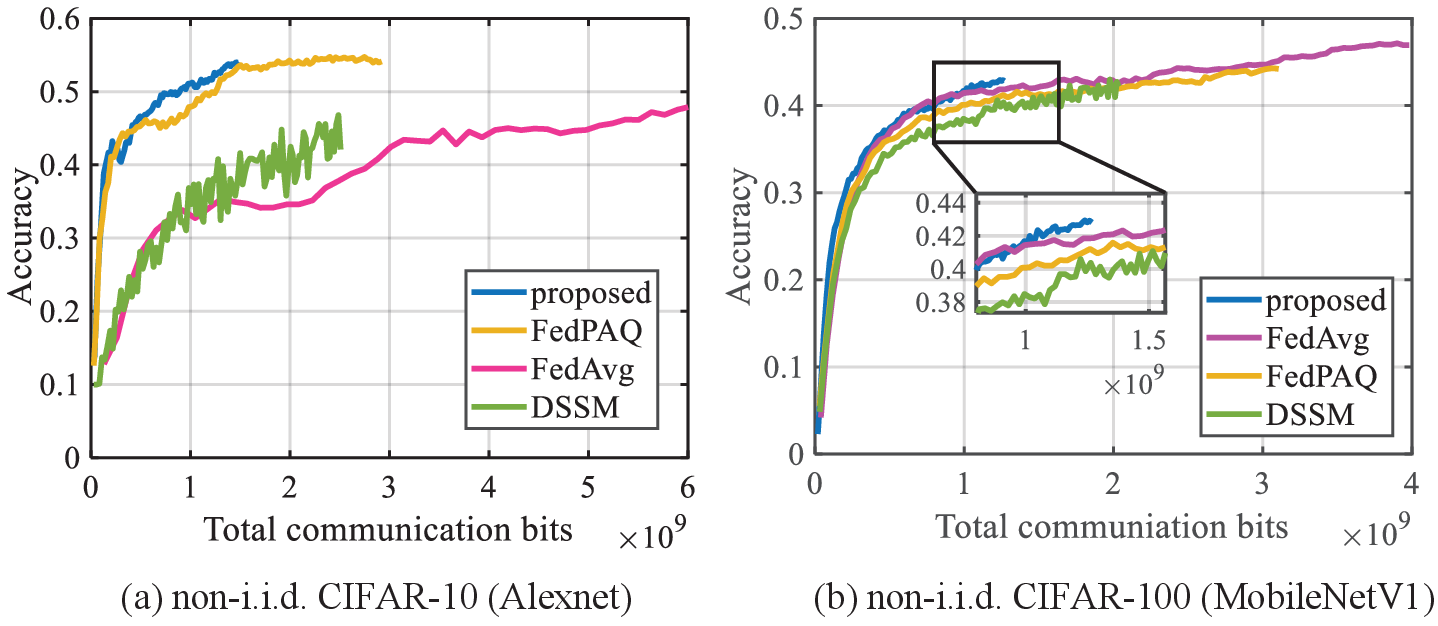}
\par\end{centering}
\caption{Accuracy versus the total communication bits in the first 100 communication
rounds.\label{fig:Accuracy-versus-bits}}
\end{figure}

In Fig. \ref{fig:Accuracy-versus-bits}, we plot the accuracy versus
total communication bits for the first 100 communication rounds. It
can be clearly observed that our proposed scheme can achieve the highest
accuracy with significantly smaller communication volume. The reason
is three-fold: 1) Our proposed scheme can achieve a highly sparse
model. This can also be observed from Table \ref{tab:Performance cmoparison lenet}.
2) The baselines can only achieve a random sparse structure, which
requires to transmit the exact location for each non-zero element.
However, in our proposed method, we only need to transmit the size
and location of each cluster. 3) Our proposed scheme can efficiently
promote the common structure among all local models, which leads to
reduced downstream communication volume. However, the existing baselines
cannot achieve downstream compression.
\begin{table}[tbh]
\begin{spacing}{0.83}
\centering{}\caption{Final model performance comparison for Alexnet.\label{tab:Performance cmoparison lenet}}
{\footnotesize{}}%
\begin{tabular}{ccccc}
\toprule 
{\footnotesize{}Task} & {\footnotesize{}Algorithm} & {\footnotesize{}Accuracy} & {\footnotesize{}Sparsity} & {\footnotesize{}Inf. time}\tabularnewline
\midrule
 & {\footnotesize{}Proposed} & \textbf{\footnotesize{}58.8\%} & \textbf{\footnotesize{}18.6\%} & \textbf{\footnotesize{}31.6ms}\tabularnewline
{\footnotesize{}CIFAR-10} & {\footnotesize{}FedPAQ} & {\footnotesize{}57.7\%} & {\footnotesize{}91.9\%} & {\footnotesize{}50.71ms}\tabularnewline
 & {\footnotesize{}DSSM} & {\footnotesize{}47.0\%} & {\footnotesize{}88.1\%} & {\footnotesize{}50.38ms}\tabularnewline
\bottomrule
\end{tabular}
\end{spacing}
\end{table}

In Table \ref{tab:Performance cmoparison lenet}, we compare the overall
performance of the final models in CIFAR-10 experiment. The sparsity
rate is measured by the percentage of non-zero weights ($\frac{\left\Vert \mathbf{w}\neq0\right\Vert _{0}}{\left\Vert \mathbf{w}\right\Vert _{0}}$),
and the inference time is measured by the GPU time of a forward pass
of 3000 samples. It can be observed that our proposed scheme can achieve
the highest accuracy in the experiment, while achieving a significantly
lower sparsity rate at inference time. The low sparse rate shows that
our proposed scheme can efficiently avoid loss of sparsity after aggregation.
While the low inference time shows that our final model has very high
computation efficiency due to the clustered structure. 
\begin{figure}[tbh]
\begin{centering}
\includegraphics[width=0.49\textwidth]{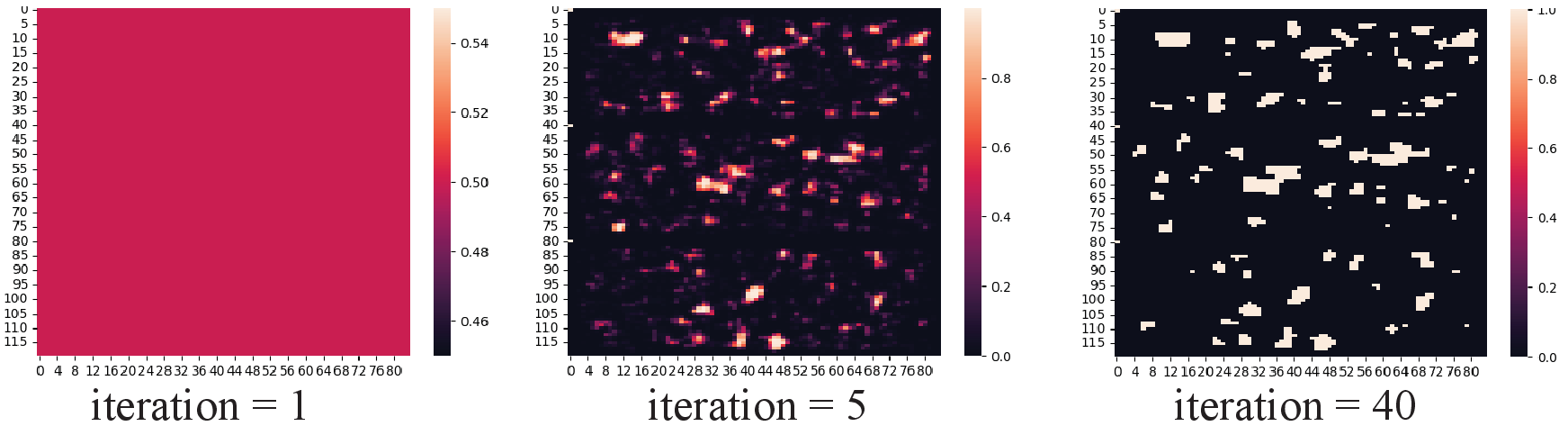}
\par\end{centering}
\caption{Message $v_{h\rightarrow s_{n}}$ of Dense\_2 layer in different iterations.\label{fig:message_structure}}
\end{figure}

In Fig. \ref{fig:message_structure}, we visualize the message $v_{h\rightarrow s_{n}}$
of Dense\_2 layer in different iterations in the CIFAR-10 experiment.
It can be clearly observe that the clustered structure is gradually
promoted during the federated iterations.

\section{Conclusion}

We proposed a Bayesian FL framework which can achieve communication
and computation efficiency simultaneously. We firstly proposed an
HMM-based hierarchical prior which can promote an novel clustered
sparse structure in the DNN weight matrix. We then proposed a D-Turbo-VBI
algorithm which can promote a common sparse structure among all the
local models. The convergence property for the proposed D-Turbo-VBI
algorithm was also established. Experiments illustrated the superior
performance of our proposed method.

\newpage{}

\bibliographystyle{IEEEtran}
\bibliography{fl_structured_model_compression_icc}

\end{document}